\pdfoutput=1

\documentclass[11pt]{article}

\usepackage[final]{ACL2023}

\usepackage{times}
\usepackage{latexsym}
\usepackage{graphicx}

\usepackage[T1]{fontenc}

\usepackage[utf8]{inputenc}

\usepackage{microtype}

\usepackage{tabularx}
\usepackage{booktabs}
\usepackage{subfigure}  
\usepackage{caption}
\usepackage{inconsolata}

\usepackage{booktabs, multirow} 
\usepackage{soul}
\usepackage{xcolor,colortbl} 
\usepackage{changepage,threeparttable} 

\title{Can Prompting LLMs Unlock Hate Speech Detection across Languages? \\
A Zero-shot and Few-shot Study}

\author{Faeze Ghorbanpour$^{1,2}$\qquad Daryna Dementieva$^{1}$\qquad Alexander Fraser$^{1, 2}$ \vspace{.2cm}\\ 
$^{1}$School of Computation, Information and Technology, TU Munich \\
$^{2}$Munich Center for Machine Learning (MCML)\\
\vspace{.1cm} {\tt \small faeze.ghorbanpour@tum.de, daryna.dementieva@tum.de} 
}

\begin{document}
\maketitle
\begin{abstract}
Despite growing interest in automated hate speech detection, most existing approaches overlook the linguistic diversity of online content.
Multilingual instruction-tuned large language models such as LLaMA, Aya, Qwen, and BloomZ offer promising capabilities across languages, but their effectiveness in identifying hate speech through zero-shot and few-shot prompting remains underexplored.
This work evaluates LLM prompting-based detection across eight non-English languages, utilizing several prompting techniques and comparing them to fine-tuned encoder models.
We show that while zero-shot and few-shot prompting lag behind fine-tuned encoder models on most of the real-world evaluation sets, they achieve better generalization on functional tests for hate speech detection.
Our study also reveals that prompt design plays a critical role, with each language often requiring customized prompting techniques to maximize performance\footnote{The code and prompts will be publicly released.}.

\end{abstract}

\section{Introduction}
Hate speech is a worldwide issue that undermines the safety of social media platforms, no matter the language \citep{thomas2021sok}. It can violate platform rules, damage user trust, influence opinions, and reinforce harmful biases against individuals or groups targeted \citep{macavaney2019hate, vedeler2019hate, stockmann2023social}. However, recent advancements in hate speech detection have been largely focused on English, as most datasets and language models are centered on English content, resulting in limited attention to other languages \citep{huang2023chatgpt, peng2023does}. Since users on social media write and engage with content in many languages—not just English—it is crucial to find tools that can detect hate speech in a multilingual context.

Instruct-tuned Large language models (LLMs) have demonstrated exceptional performance across a wide range of text-related tasks \citep{skibicki2025llms, zhang2023instruction}. Many of these models possess multilingual capabilities, enabling them to process and understand text in various languages \citep{pedrazzini2025multilingual, shaham-etal-2024-multilingual}. This makes them suitable for tasks like hate speech detection without the need for additional fine-tuning, thereby reducing computational and resource costs. While their effectiveness in detecting hate speech in English has been studied extensively \citep{roy2023probing, guo2023investigation, zhang2025llm}, their performance on non-English datasets remains underexplored.

To evaluate the capabilities of multilingual instruction-tuned LLMs in detecting hate speech in various languages, we conduct a study using several prompting techniques, including zero-shot prompting (e.g., vanilla, chain-of-thought, role-play), few-shot prompting, and combinations of these prompts. We evaluate performance across eight non-English hate speech detection tasks, covering Spanish, Portuguese, German, French, Italian, Turkish, Hindi, and Arabic, using real-world\footnote{By real-world test sets, we meant datasets collected from actual conversations, which better reflect real-world scenarios.} and hate speech functional test sets. This study seeks to address the following research questions:
(1) How well do LLMs perform on hate speech detection across various non-English languages?
(2) Does few-shot prompting improve performance compared to zero-shot prompting?
(3) How does LLM performance compare to that of traditional fine-tuned models?

Our findings highlight the importance of prompt design in multilingual hate speech detection. While performance varies by the prompting strategy, experimenting with different techniques leads to reasonably strong results. In most languages, few-shot prompting combined with other techniques outperforms zero-shot prompting, suggesting that providing a few task-specific examples is beneficial.


Compared to fine-tuned encoder models, prompting LLMs shows lower performance on real-world test sets. However, in functional test cases, prompting often performs better. Further analysis of languages where prompting underperforms on real-world data suggests that prompting can still be a practical option when only limited training data is available. Nonetheless, with access to larger training sets, fine-tuning encoder models remains the more effective approach. Overall, instruction-tuned LLMs demonstrate stronger generalization in controlled functional benchmarks, without the need for additional training.

\section{Related Work}
The ability of instruction-tuned LLMs to perform a wide range of NLP tasks without the need for fine-tuning or training data has drawn growing interest, particularly in applications like hate speech detection. 
Recent studies have explored LLM-based hate speech detection, primarily in English. \citet{zhu2023can} reports low agreement between LLM predictions and human annotations, while \citet{li2024hot} finds that LLMs are more effective at identifying non-hateful content. 
\citet{huang2023chatgpt} examines the use of LLMs for generating explanations of implicit hate, and \citet{roy2023probing} shows that including target-specific information in prompts improves performance.

Another study examines how in-context learning, combined with few-shot examples and task descriptions, boosts the performance of hate speech detection by LLMs \citep{han2022designing}. \citet{guo2023investigation} investigates using LLMs for real-world hate speech detection using four diverse prompting strategies and finds that few-shot and chain-of-thought prompts help.
While these works have explored prompting techniques, they primarily assess the capabilities of LLMs for hate speech detection in English and do not examine a broad range of prompting strategies across languages.

There have been efforts to investigate the capabilities of LLMs for non-English hate speech. \citet{guo2023investigation} and \citet{math12233687} tested prompt strategies only in Chinese and Bangla, respectively. \citet{masud2024hate} assesses LLMs' sensitivity to geographical priming and persona attributes in five languages, showing that geographical cues can improve regional alignment in hate speech detection. Similarly, \citet{zahid2025evaluation} uses geographical contextualization into prompts for five languages. These motivate our use of language-aware prompts; However, they do not explore a wide range of prompting strategies, such as few-shot, chain-of-thought, or role-play prompts.

\citet{tonneau2024hateday} evaluate hate speech detection in eight languages using real-world and functional test sets, but rely solely on vanilla prompting.
\citep{dey2024better} applied prompting LLMs to three low-resource South Asian languages, finding that translating inputs to English outperformed prompting in the original language. This motivated us to prompt the LLM to translate before classifying. In contrast to these efforts, our work covers eight languages and evaluates a broader range of prompt designs on real-world and functional test sets.

\section{Datasets} 
We selected datasets with explicit hate speech labels that follow a definition consistent with our story: \textit{abusive language that targets a protected group or individuals for being part of that group} \citep{rottger2020hatecheck}.
We randomly selected \underline{2,000 samples} from each dataset as test sets for evaluating both prompting and fine-tuned models. For Arabic and French, smaller dataset sizes limited the test sets to 1,000 and 1,500 samples, respectively. The remaining data was used to train the encoder models.

The datasets are as follows:
\textbf{OUS19\_AR} \citep{ousidhoum2019multilingual}: Contains 3,353 Arabic tweets, with 22.5\% labeled as hateful. \textbf{OUS19\_FR} \citep{ousidhoum2019multilingual}: Consist of 4,014 French tweets, with 11.0\% labeled as hateful.
 \textbf{BAS19\_ES} \citep{basile2019semeval}: Compiled for SemEval 2019, it includes 4,950 Spanish tweets, 41.5\% of which are labeled as hateful.
 \textbf{HAS21\_HI} \citep{modha2021overview}: Collected for HASOC 2021, it contains 4,594 Hindi tweets, with 12.3\% labeled hateful.
 \textbf{SAN20\_IT} \citep{sanguinetti2020haspeede}: Created for Evalita 2020, it includes 8,100 Italian tweets, 41.8\% of which are hateful.
 \textbf{FOR19\_PT} \citep{fortuna2019hierarchically}: Consists of 5,670 Portuguese tweets, with 31.5\% labeled as hateful.
 \textbf{Gahd24\_DE} \citep{goldzycher2024improving}: A German adversarial dataset consisting of 10,996 tweets, 42.4\% of which are labeled as hateful.
 \textbf{Xdomain\_TR} \citep{toraman-etal-2022-large}: A large-scale, multi-domain Turkish dataset consisting of 38K samples, with a class imbalance rate of 74.4\%.

For functional hate speech evaluation, we used the \textbf{HateCheck benchmark} \citep{rottger2020hatecheck}, a benchmark for evaluating the robustness of hate speech detection systems across languages. It includes functional test cases—controlled examples designed to test specific capabilities such as handling implicit hate, negation, and non-hateful slurs. Originally developed for English, it has been extended by \citet{rottger2022multilingual} to multiple languages to support cross-lingual evaluation and reveal systematic model weaknesses not captured by standard datasets.

\section{Models}
We evaluate four instruction-tuned multilingual LLMs for hate speech detection across eight languages:
\textbf{LLaMA-3.1-8B-Instruct} \citep{grattafiori2024llama}: Meta’s instruction-tuned decoder model, optimized for reasoning tasks and primarily designed for English, with multilingual support.
\textbf{Qwen2.5-7B-Instruct} \citep{qwen2.5, qwen2}: A multilingual decoder model by Alibaba Cloud, supporting 30+ languages with strong instruction-following capabilities.
\textbf{Aya-101} \citep{ustun-etal-2024-aya}: Cohere’s multilingual model trained on 100+ languages, tuned for equitable cross-lingual NLP, including hate speech detection.
\textbf{BloomZ-7B1} \citep{muennighoff2022crosslingual}: A decoder model by BigScience, fine-tuned via multitask instruction tuning on 46 languages for cross-lingual instruction following.

For the encoder-based baseline, we fine-tuned two multilingual models with strong performance on classification tasks:
\textbf{XLM-T} \citep{barbieri2022xlmt, conneau2020unsupervised}: An XLM-R extension pre-trained on 198M Twitter posts in 30+ languages.
\textbf{mDeBERTa} \citep{he2021deberta}: A multilingual encoder covering 100+ languages, effective in zero-shot and low-resource settings.
Further implementation details and hyperparameters are provided in Appendix~\ref{sec:model_details}.



\section{Prompts}
We assess instruction-tuned multilingual LLMs using a range of prompting strategies for hate speech detection, such as: directly asking whether a comment is hateful (vanilla); prompting the model to act as a classifier (classification); chain-of-thought prompting for step-by-step reasoning (CoT); natural language inference-inspired (NLI) prompts; language-aware prompts that consider linguistic and cultural context (multilingual); assigning the LLM the role of a community moderator (role-play); translate then classify prompts (translate); definition-based prompts that explain hate speech (definition); and defining related forms of abusive content to help the model differentiate them from hate speech (distinction), etc. We also include few-shot prompting, where we retrieve and insert example instances from the training set into the prompt. We also explore combinations of these strategies. For full prompt texts and implementation details, see Appendix~\ref{sec:prompt_details}.

\begin{table*}[!htp]\centering
\scriptsize
\begin{tabular}{p{0.05cm}p{0.1cm}p{1cm}p{0.3cm}p{1cm}p{0.3cm}p{1cm}p{0.3cm}p{2cm}p{0.3cm}p{1.6cm}p{0.3cm}p{1.8cm}p{0.3cm}}\toprule
& &\multicolumn{2}{c}{BloomZ} &\multicolumn{2}{c}{Aya101} &\multicolumn{4}{c}{Llama3} &\multicolumn{4}{c}{Qwan} \\\cmidrule(rl){3-4}\cmidrule(rl){5-6}\cmidrule(rl){7-10}\cmidrule(rl){11-14}

& &\multicolumn{2}{c}{zero-shot} &\multicolumn{2}{c}{zero-shot} &\multicolumn{2}{c}{zero-shot} &\multicolumn{2}{c}{few-shot} &\multicolumn{2}{c}{zero-shot} &\multicolumn{2}{c}{few-shot} \\\cmidrule(rl){3-4}\cmidrule(rl){5-6}\cmidrule(rl){7-8}\cmidrule(rl){9-10}\cmidrule(rl){11-12}\cmidrule(rl){13-14}

& &prompt &f1 &prompt &f1 &prompt &f1 &prompt &f1 &prompt &f1 &prompt &f1 \\\midrule
\multirow{8}{*}{\rotatebox{90}{Real World Tests}} &es &classification &54.50 &definition &62.68 &classification &63.13 &5 shot + CoT &\textbf{68.89} &translate &64.79 &5 shot + CoT &\textbf{68.90} \\
&pt &definition &63.92 &definition &71.51 &role-play &70.79 &5 shot + multilingual &\textbf{73.7} &role-play + CoT &\textbf{73.44} &5 shot + role-play &72.56 \\
&hi &multilingual &51.33 &classification &47.33 &CoT &52.09 &1 shot + CoT &\textbf{54.90} &distinction &53.76 &1 shot + CoT &49.57 \\
&ar &NLI &58.67 &distinction &64.67 &classification &62.66 &3 shot + definition &64.71 &NLI &\textbf{70.61} &3 shot &67.36 \\
&fr &NLI &55.63 &translate &53.44 &CoT &55.22 &5 shot + definition &51.53 &NLI &\textbf{55.59} &3 shot &52.22 \\
&it &CoT &55.50 &vanilla &74.82 &distiction &75.86 &5 shot + CoT &76.18 &multilingual &73.34 &5 shot + CoT &\textbf{76.57} \\
&de &CoT &38.36 &general &67.51 &role-play &50.16 &1 shot + multilingual &75.16 &target &50.19 &5 shot + definition &\textbf{77.55} \\
&tr &role-play &55.20 &- &- &classification &76.16 &5 shot + CoT &\textbf{81.76}&translate &75.89 &5 shot + CoT &77.03 \\\midrule
\multirow{7}{*}{\rotatebox{90}{Functional Tests}} &es &definition &64.88 &distinction &73.19 &vanilla &86.37 &5 shot &\textbf{87.40} &vanilla &84.39 &5 shot + definition &86.43 \\
&pt &definition &66.04 &distinction &72.39 &classification &\textbf{83.37} &3 shot &\textbf{86.59} &CoT &82.15 &5 shot + definition &84.08 \\
&hi &role-play &51.99 &distinction &\textbf{65.95} &classification &65.31 &1 shot + multilingual &65.36 &definition &65.41 &1 shot + definition &\textbf{65.93} \\
&ar &definition &62.08 &vanilla &62.99 &contextual &64.00 &1 shot &67.95 &general &70.42 &3 shot + definition &\textbf{71.88} \\
&fr &CoT &62.62 &distinction &71.94 &vanilla &\textbf{84.61} &5 shot + role-play &\textbf{84.37} &vanilla &82.06 &5 shot + role-play &82.61 \\
&it &role-play &55.15 &distinction &71.25 &role-play &78.54 &5 shot &82.95 &target &78.35 &5 shot + definition &\textbf{84.17} \\
&de &role-play &51.75 &distinction &72.64 &contextual &83.27 &5 shot + multilingual &\textbf{89.65} &contextual &82.64 &5 shot + definition &86.62 \\
\bottomrule
\end{tabular}
\caption{Zero-shot and Few-shot Prompting Results for Instruction-tuned Multilingual LLMs.(f1=f1-macro)}\label{tab:promptin_results}
\end{table*}

\section{Results}
We evaluate instruction-tuned LLMs with various prompt types over three runs in the inference mode and report the average F1-macro scores. Table~\ref{tab:promptin_results} summarizes the performance of zero-/few-shot results for four instruction-tuned models across eight languages.
We observe that prompt design significantly affects performance. \textit{Aya101} performs best with definition- and distinction-based prompts, suggesting that explicit definitions improve its accuracy. In contrast, \textit{Qwen} excels with NLI and role-play prompts, indicating sensitivity to context and conversational cues.

In zero-shot settings, Qwen and LLaMA3 generally outperform the other models, with similar overall performance. However, Qwen performs better in most real-world test cases, whereas LLaMA3 leads on functional benchmarks. Few-shot prompting (typically five-shot) improves performance, especially on functional tests, as examples help the model apply contextual distinctions more effectively. On real-world tests, improvement is less consistent—even with examples from the same training data. This suggests that few-shot effectiveness depends not only on data quality, but also on prompt clarity and structure.
Overall, instruction-tuned LLMs perform notably well on functional tests and reasonably on real-world tests in different languages. However, their effectiveness depends heavily on prompt design and the inclusion of few-shot examples.
Appendix~\ref{sec:full_prompting_results} contains detailed performance results for Spanish and Portuguese.

\begin{figure*}[htbp]
    \centering
    \subfigure{\includegraphics[width=0.325\textwidth]{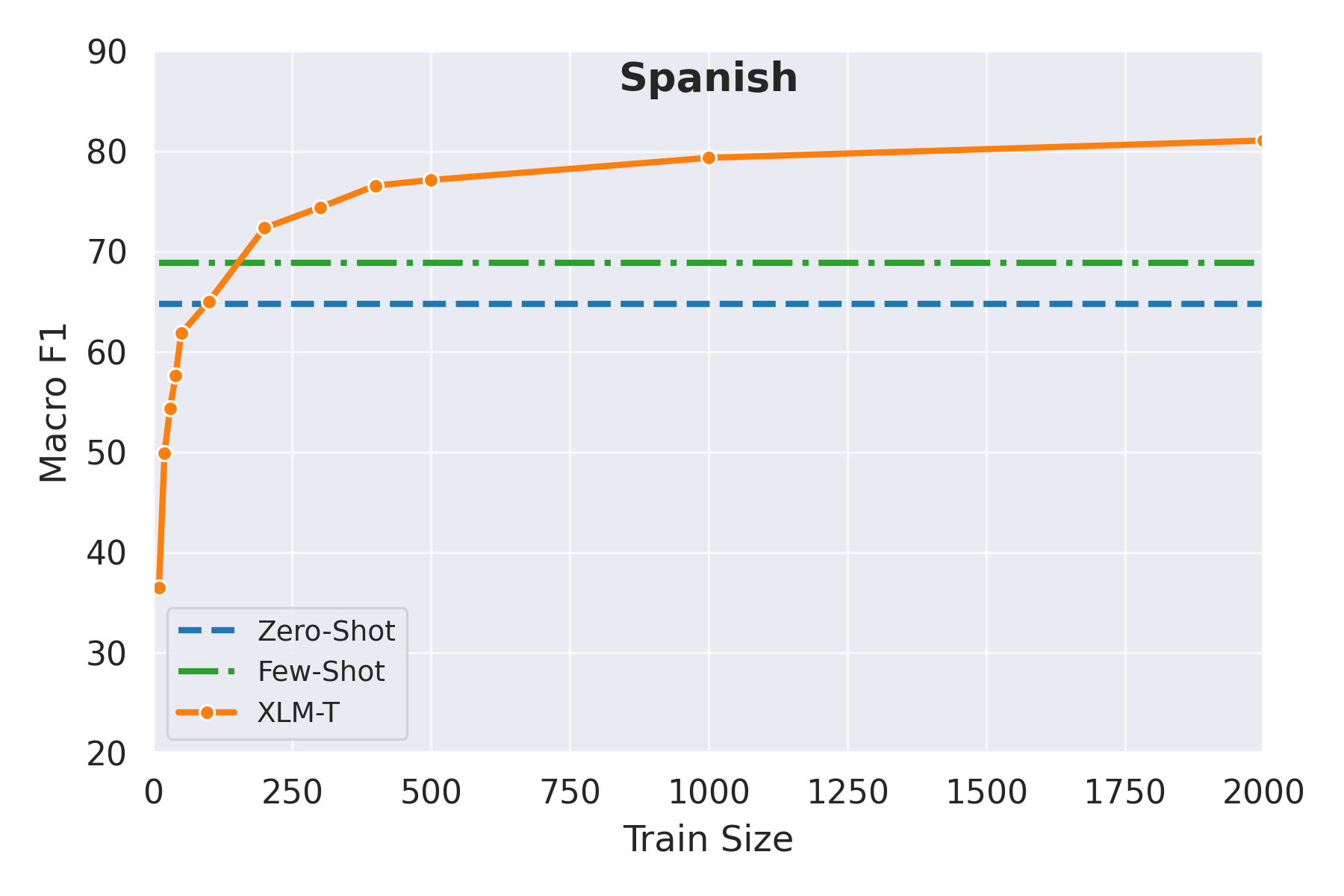}}
    \subfigure{\includegraphics[width=0.325\textwidth]{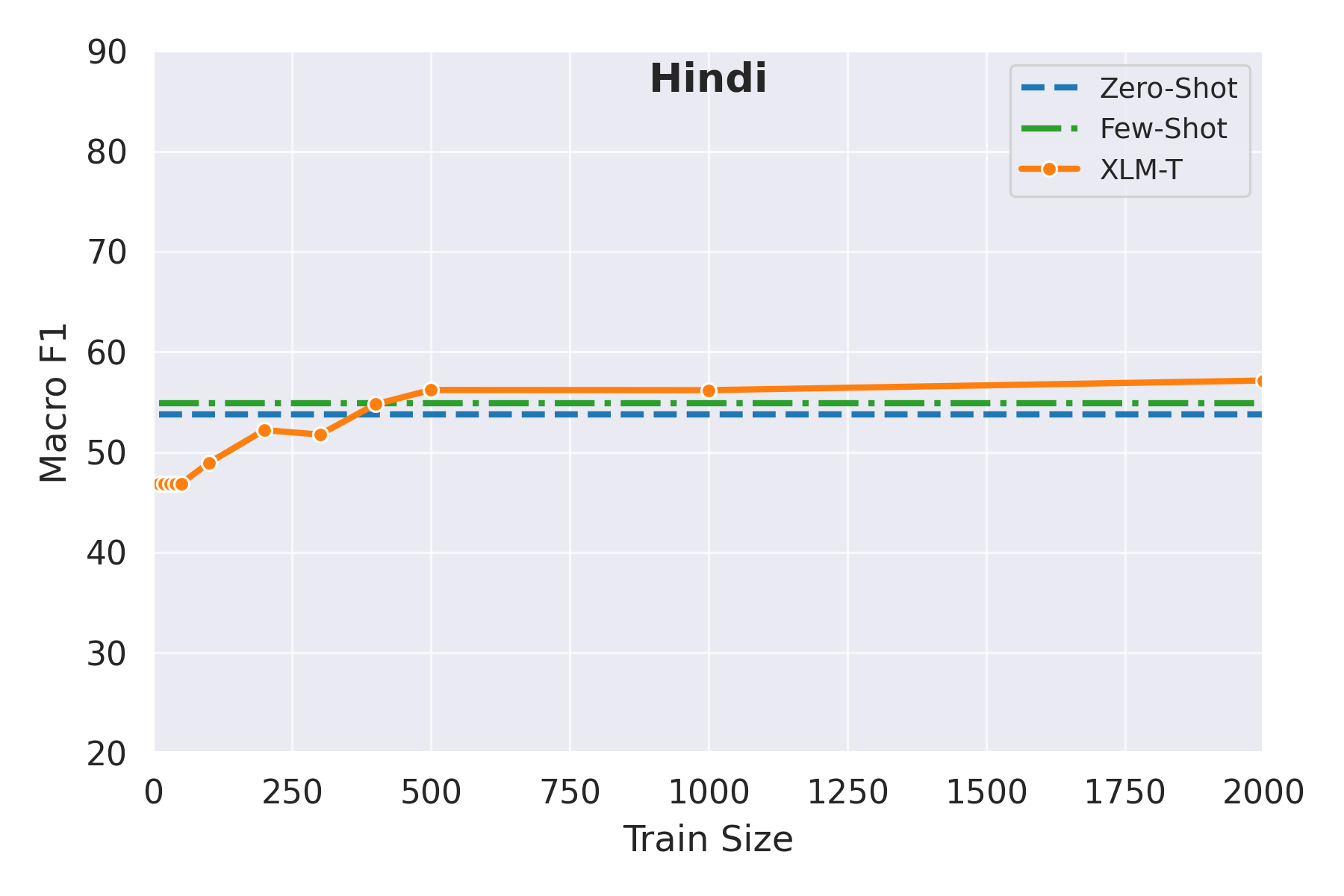}}
    \subfigure{\includegraphics[width=0.325\textwidth]{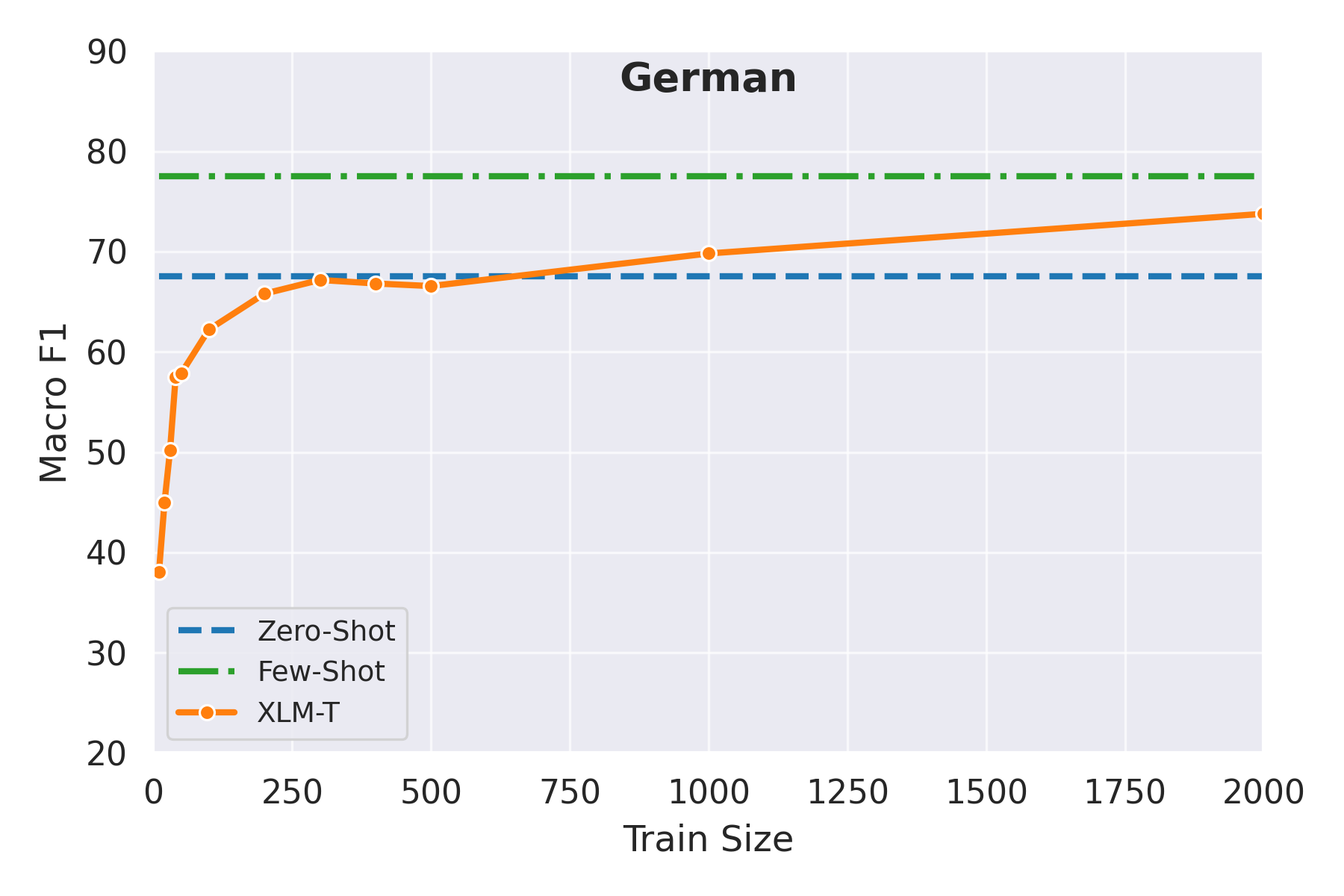}}
    \captionsetup{skip=3pt}  
    \caption{Performance of zero-/few-shot prompted LLMs vs. fine-tuned XLM-T across varying training sizes.}
    \label{fig:multi_images}
\end{figure*}

For comparison, we fine-tune two encoder models for binary hate speech classification on train sets of datasets using five random seeds and report the average macro F1 scores. Table~\ref{tab:prompting_vs_finetuning} summarizes the performance of encoder models alongside the best zero- and few-shot prompting results. On real-world datasets, encoder models generally outperform LLM prompting across most languages, benefiting from fine-tuning on task-specific data. However, the trend reverses on functional tests, where few-shot prompting often yields better results—highlighting the stronger generalization ability of large LLMs in controlled evaluation settings.

\begin{table}[!htp]\centering
\scriptsize
\begin{tabular}{p{0.1cm}p{0.2cm}p{1.2cm}p{1.2cm}p{1.2cm}p{1.2cm}r}\toprule
& &fine-tuned mDeBERTa &fine-tuned XLM-T &zero-shot prompting &few-shot prompting \\\midrule
\multirow{8}{*}{\rotatebox{90}{Real World Tests}} &es &81.45 &\textbf{82.78} &64.79 &68.90 \\
&pt &\textbf{73.22} &72.62 &73.44 &\textbf{73.70} \\
&hi &51.34 &\textbf{59.18} &53.76 &54.90 \\
&ar &68.34 &\textbf{70.31} &\textbf{70.61} &67.36 \\
&fr &51.56 &51.42 &\textbf{55.59} &52.22 \\
&it &\textbf{79.71} &78.82 &75.86 &76.57 \\
&de &\textbf{80.39} &79.18 &67.51 &77.55 \\
&tr &\textbf{92.72} &88.32 &76.16 &81.76 \\\midrule
\multirow{7}{*}{\rotatebox{90}{Functional Tests}} &es &60.94 &67.93 &86.37 &\textbf{87.40} \\
&pt &58.94 &57.28 &83.37 &\textbf{86.59} \\
&hi &24.91 &23.26 &\textbf{65.95} &\textbf{65.93} \\
&ar &23.93 &25.47 &70.42 &\textbf{71.88} \\
&fr &25.89 &26.61 &\textbf{84.61} &\textbf{84.37} \\
&it &54.07 &52.05 &78.54 &\textbf{84.17} \\
&de &74.36 &70.60 &83.27 &\textbf{89.65} \\
\bottomrule
\end{tabular}
\caption{Results (f1-macro) of fine-tuned encoder models vs. best zero-/few-shot prompting LLMs.}\label{tab:prompting_vs_finetuning}
\end{table}

To understand when prompting is preferable, we conducted additional experiments comparing encoder model performance at varying training set sizes to that of prompting. Figure~\ref{fig:multi_images} presents results for three languages where prompting underperforms compared to fine-tuned models. Depending on the language, prompting becomes competitive when training data is limited—for example, with 100–200 examples in Spanish, 300–400 in Hindi, or 600-700 in German. Beyond that, fine-tuning generally yields better performance. See Appendix~\ref{sec:fine_tuning_plots} for more results across other languages.

\section{Conclusion}
In this study, we explore the capabilities of multilingual instruction-tuned LLMs in detecting hate speech across eight non-English languages. The findings suggest that different prompting techniques work better for different languages, indicating that it is beneficial to experiment with various prompt designs when addressing a new language. In real-world scenarios, where the data is more culturally dependent, prompting LLMs is less effective than training encoder models with task-specific data. However, in functional hate speech tests, LLMs tend to perform better and offer more flexibility. Incorporating few-shot examples into prompts in such cases may further enhance the LLMs' performance.

\section*{Limitations}
One unavoidable limitation of our work is the number of multilingual instruction-tuned LLMs we were able to include. Given the rapid growth and proliferation of generative AI models, new LLMs are continually emerging. However, due to resource and time constraints, we were unable to include more models in our evaluation.

A second limitation concerns the additional contextual information available for prompt construction. Most of our datasets included only the text, label, and language, but lacked richer metadata. Incorporating information such as the targeted group of the hate speech, the context in which it occurred, or the domain of the text could potentially improve model performance \citep{roy2023probing}.

Moreover, we treated these LLMs as black-box models and did not attempt to analyze their internal parameters or architectural components. We also did not fine-tune the larger models to better adapt them to our datasets.

Finally, while we incorporated a wide range of carefully designed prompt variations to probe model behavior, our set of prompt configurations is not exhaustive. There may exist alternative formulations or edge cases that we have not explored. Therefore, our findings should be interpreted as indicative rather than definitive.


\section*{Acknowledgements}
The work was supported by the European Research Council (ERC) through the European Union's Horizon Europe research and innovation programme (grant agreement No. 101113091) and the German Research Foundation (DFG; grant FR 2829/7-1).

\bibliography{custom}
\bibliographystyle{acl_natbib}

\onecolumn
\appendix

\section{Model and Training Details}
\label{sec:model_details}

\subsection{LLM Selection and Setup}
To select suitable instruction-tuned multilingual LLMs, we first conducted a brief experiment to ensure that their safety tuning would not interfere with hate speech classification. Our goal was to evaluate detection capabilities, not robustness to jailbreak attempts. We excluded models such as mT0-large\citep{muennighoff2023crosslingualgeneralizationmultitaskfinetuning}, Ministral-8B-Instruct\citep{mistral2024ministraux}, and Teuken-7B-instruct \citep{ali2024teuken7bbaseteuken7binstructeuropean} that failed to follow instructions reliably.
We used the \texttt{transformers} library\footnote{\url{https://huggingface.co/docs/transformers/en/index}} to load and run models in inference mode, generating binary outputs (\texttt{yes} or \texttt{no}). We set \texttt{max\_new\_tokens=10}, \texttt{do\_sample=False}, and left temperature/top-k/top-p unset. Batch size and max sequence length varied depending on the prompt and model. Each experiment was run with three random seeds, and we also swapped the position of \texttt{yes} and \texttt{no} in prompts to mitigate position bias.

\subsection{Encoder Model Training}
For training the encoder-based models, in addition to the previously mentioned 2,000-sample test set, we randomly held out 500 samples for validation and used the remaining data for training. Models were fine-tuned for 10 epochs using the \texttt{transformers} \texttt{Trainer}, with a batch size of 16 and max sequence length of 128. Default settings were used for the learning rate, optimizer, and scheduler. We evaluated the models on both the test set of the corresponding dataset and its language-specific subset in the HateCheck benchmark.

\subsection{Data Formatting}
Most datasets used were binary hate vs. non-hate classification tasks. Any remaining datasets such as German and Turkish ones were also converted to this binary format to ensure consistency. Most of the datasets used in this study are sourced from the GitHub repository by \citet{rottger2022data}:\url{https://github.com/paul-rottger/efficient-low-resource-hate-detection}. The German dataset is available at:\url{https://github.com/jagol/gahd}, and the Turkish dataset is from:\url{https://github.com/avaapm/hatespeech}. These datasets are legally licensed and permitted for use in research projects.

\subsection{Model Size and Budget}
Experiments with instruction-tuned LLMs—LLaMA3\footnote{\url{https://huggingface.co/meta-llama/Llama-3.1-8B-Instruct}}, Qwen2.5\footnote{\url{https://huggingface.co/Qwen/Qwen2.5-7B-Instruct}}, Aya101\footnote{\url{https://huggingface.co/CohereLabs/aya-101}}, and BloomZ\footnote{\url{https://huggingface.co/bigscience/bloomz-7b1}}—were primarily conducted on NVIDIA RTX A6000 servers in inference mode, with no parameter updates during prompting. In contrast, fine-tuning of encoder models was performed on NVIDIA GeForce GTX 1080 Ti GPUs, where all model parameters were updated during training. The mDeBERTa\footnote{\url{https://huggingface.co/microsoft/mdeberta-v3-base}} has approximately 86 million parameters, while XLM-T\footnote{\url{https://huggingface.co/cardiffnlp/twitter-xlm-roberta-base}} consists of around 279 million parameters. All models used in this study were sourced from Hugging Face and are licensed for legal use in academic research.

\section{Prompts Details}
\label{sec:prompt_details}
You can find the zero-shot prompt texts in Table~\ref{tab:zero_shot_prompt_table_1}and Table~\ref{tab:zero_shot_prompt_table_2} and the few-shot prompt texts in Table~\ref{tab:few_shot_prompt_table}. In these tables, "CoT" refers to chain-of-thought prompting, and "NLI" refers to prompts guided by natural language inference. The "+" symbol indicates a combination of the two prompt types.
Dataset instances are enclosed in \{\texttt{text}\}. The placeholders \{\texttt{yn1}\} and \{\texttt{yn2}\} correspond to the expected outputs ("yes" and "no"), which were randomly swapped to reduce positional bias. The input language is represented as \{\texttt{language}\}, and if the prompt requires few-shot examples, they are inserted in \{\texttt{examples}\}.

In our few-shot experiments, we retrieve 1, 3, or 5 examples per class from the training set and include them in the prompt to guide the model's predictions. For example, a 5-shot setting includes 5 hate and 5 non-hate examples, interleaved by class (e.g., one hate, one non-hate), resulting in a total of 10 examples. While we experimented with using more than 5 examples per class, context length limitations and computational constraints prevented us from applying this to all prompts and models.

\begin{table*}[htbp]
\footnotesize
\centering
\begin{tabularx}{\linewidth}{p{2.5cm} X}
\toprule
\textbf{Prompt Type} & \textbf{Prompt Text} \\
\midrule

vanilla & Is the following comment hateful? Respond only with \{yn1\} or \{yn2\}. Don't provide explanations or any other information.\\
& Comment: "\{text\}". Answer: \\

\addlinespace

classification & Classify the following comment as either hateful or non-hateful. Respond with yes for hateful and no for non-hateful. Answer only \{yn1\} or \{yn2\}. Don't provide explanations or any other information.\\
& Comment: "\{text\}". Answer: \\
\addlinespace

definition & Hate speech is language that expresses discrimination, hostility, or violence against individuals or groups based on race, religion, ethnicity, nationality, gender, sexual orientation, or disability.
Does the following comment contain hate speech? Answer only \{yn1\} or \{yn2\}. Don't provide explanations or any other information.\\
& Comment: "\{text\}". Answer: \\
\addlinespace

CoT & Analyze the following comment step by step to determine if it contains hate speech.\\
&Step 1: Identify if the comment targets a specific person or group based on race, religion, ethnic origin, national origin, sex, disability, sexual orientation, or gender identity. If no target is identified, the comment is not hate speech.\\
&Step 2: Assess whether the language expresses hostility, discrimination, incitement to violence, dehumanization, or derogatory stereotypes toward the target.\\
&Step 3: Based on the above steps, conclude whether this is hate speech. \\
&Answer only \{yn1\} or \{yn2\}. Don't provide explanations or any other information.\\
& Comment: "\{text\}". Answer: \\
\addlinespace

NLI & Premise: "\{text\}"\\
&Hypothesis: "This statement contains hate speech as defined by social and legal standards."\\
&Does the premise entail the hypothesis? Don't provide explanations or any other information. Answer only \{yn1\} or \{yn2\}. Answer:\\
\addlinespace

role-play & You are a community moderator responsible for ensuring compliance with hate speech policies.
Analyze the following text and determine if it violates the guidelines.\\
&Text: "\{text\}"\\
&Does this text violate the community’s hate speech policy? Don't provide explanations or any other information. Answer only \{yn1\} or \{yn2\}. Answer:\\
\addlinespace

multilingual + definition & Hate speech is defined as language that expresses discrimination, hostility, or violence against individuals or groups based on race, religion, ethnicity, nationality, gender, sexual orientation, or disability.\\
&Analyze the following text written in \{language\} considering cultural context, idioms, and societal norms of \{language\}-speaking communities.
Does this text contain hate speech? \\
&Don't provide explanations or any other information. Answer only \{yn1\} or \{yn2\}.\\
&Text: "\{text\}". Answer:\\
\addlinespace

role-play + cot & You are a community moderator responsible for enforcing hate speech policies. Analyze the following comment step by step to determine if it violates these policies. \\
&Step 1: Identify any offensive or harmful language targeting a person or group.\\
&Step 2: Assess if the language attacks based on attributes such as race, religion, ethnic origin, national origin, sex, disability, sexual orientation, or gender identity.\\
&Step 3: Determine if the comment violates the community’s hate speech policy.\\
&Does this comment violate the community guidelines on hate speech? Answer only \{yn1\} or \{yn2\}.
Don't provide explanations or any other information.\\
&Comment: "\{text\}". Answer: \\
\addlinespace

multilingual + cot & Analyze the following text written in \{language\} step by step, considering cultural context, idioms, and societal norms in \{language\}-speaking communities.\\
&Step 1: Identify any offensive or harmful language targeting a person or group.\\
&Step 2: Determine if the language attacks based on attributes such as race, religion, ethnic origin, national origin, sex, disability, sexual orientation, or gender identity.\\
&Step 3: Conclude whether it is hate speech.\\
&Does this text contain hate speech? 
Don't provide explanations or any other information. Answer only \{yn1\} or \{yn2\}.\\
&Text: "\{text\}". Answer: \\
\addlinespace

target & Identify the target group, if any, in the following comment. Then, determine if it contains hate speech. \\
& Does it contain hate speech? Answer only \{yn1\} or \{yn2\}.\\
&Comment: "\{text\}". Answer:\\
\bottomrule
\end{tabularx}
\caption{Zero-shot prompt templates for hate speech detection in eight non-English languages (part 1).}
\label{tab:zero_shot_prompt_table_1}
\end{table*}

\begin{table*}[htbp]
\footnotesize
\centering
\begin{tabularx}{\linewidth}{p{2.5cm} X}
\toprule
\textbf{Prompt Type} & \textbf{Prompt Text} \\
\midrule
contextual& Consider the context of the following comment, including the target audience and potential impact. Does this comment contain hate speech? \\
&Answer only \{yn1\} or \{yn2\}. Don't provide explanations or any other information.\\
&Comment: "\{text\}". Answer:\\
\addlinespace

translate & Translate the following text from \{language\} to English: Text: "\{text\}"\\
&Now, classify the text as either hateful or non-hateful. Don't provide explanations or any other information. \\
&Respond with yes for hateful and no for non-hateful. Answer: \\
\addlinespace

distinction & Considering the following definitions, tell me if the given comment contains hate speech? \\
&- Hate Speech: Attacks, dehumanizes, incites violence, or promotes discrimination against a group based on identity (e.g., race, religion, gender, etc.).\\
&- Toxic Language: Aggressive, harassing, or harmful speech, but not necessarily identity-based.\\
&- Offensive Language: Rude or impolite but does not incite harm or target a group.\\
&Answer only \{yn1\} or \{yn2\}. Don't provide explanations or any other information.\\
&Comment: "\{text\}". Answer: \\

\bottomrule
\end{tabularx}
\caption{Zero-shot prompt templates for hate speech detection in eight non-English languages (part 2).}
\label{tab:zero_shot_prompt_table_2}
\end{table*}

\begin{table*}[htbp]
\footnotesize
\centering
\begin{tabularx}{\linewidth}{p{2.5cm} X}
\toprule
\textbf{Prompt Type} & \textbf{Prompt Text} \\
\midrule
few-shot& Consider following examples:\\
&\{examples\}\\
&Now, classify this comment as either hateful or non-hateful:\\
&Comment: "\{text\}"\\
&Respond with yes for hateful and no for non-hateful. Answer only \{yn1\} or \{yn2\}. Don't provide explanations or any other information. 
Answer:\\
\addlinespace

few-shot + CoT & Here are some examples of comments and their classifications as yes if it is hateful or no if not:\\
&\{examples\}\\
&Now, analyze the following comment step by step and determine if it is hateful.\\
&Step 1: Identify any offensive or harmful language targeting a person or group.\\
&Step 2: Determine if the language attacks based on attributes such as race, religion, ethnic origin, national origin, sex, disability, sexual orientation, or gender identity.\\
&Step 3: Conclude whether it is hate speech. Don't provide explanations or any other information. \\
&Answer only \{yn1\} or \{yn2\}. Don't provide explanations or any other information.\\
&Comment: "\{text\}". Answer: \\
\addlinespace

few-shot + role-play& As a community moderator, you have previously handled the following cases if they include hate speech or not:\\
&\{examples\}\\
&Based on these examples, analyze the following comment and determine if it violates the community’s hate speech policy.\\
&Don't provide explanations or any other information. Answer only \{yn1\} or \{yn2\}.\\
&Comment: "\{text\}". Answer:\\
\addlinespace

few-shot + multilingual & Here are examples of comments in \{language\} and their classifications as yes if it is hateful or no if not:\\
&\{examples\}\\
&Now, analyze the following comment written in \{language\} and determine if it is hateful. Is this comment hateful?\\
&Don't provide explanations or any other information. Answer only \{yn1\} or \{yn2\}.\\
&Comment: "\{text\}". Answer:\\
\addlinespace

few-shot + definition & Hate speech is defined as language that expresses discrimination, hostility, or violence against individuals or groups based on race, religion, ethnicity, nationality, gender, sexual orientation, or disability. Here are some examples:\\
&\{examples\}\\
&Is this comment hateful? Comment: "\{text\}"\\
&Don't provide explanations or any other information. Answer only \{yn1\} or \{yn2\}. Answer:\\

\bottomrule
\end{tabularx}
\caption{Few-shot prompt templates for hate speech detection in non-English languages.}
\label{tab:few_shot_prompt_table}
\end{table*}

\section{Comparing Prompting and Fine-tuning Under Varying Data Conditions}
\label{sec:fine_tuning_plots}
\begin{figure*}[htbp]
    \centering
    \subfigure[Spanish]{\includegraphics[width=0.23\textwidth]{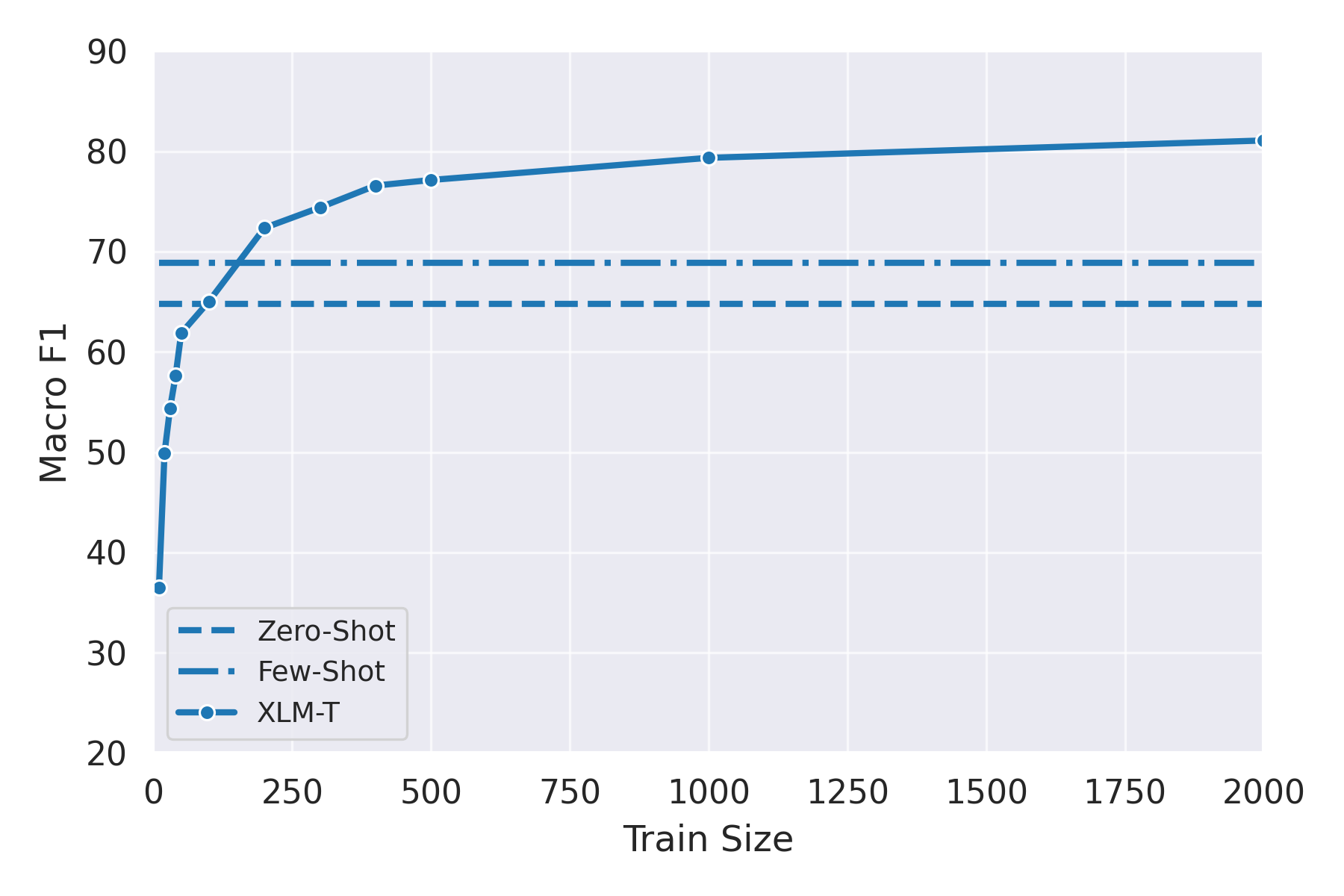}}
    \subfigure[Portuguese]{\includegraphics[width=0.23\textwidth]{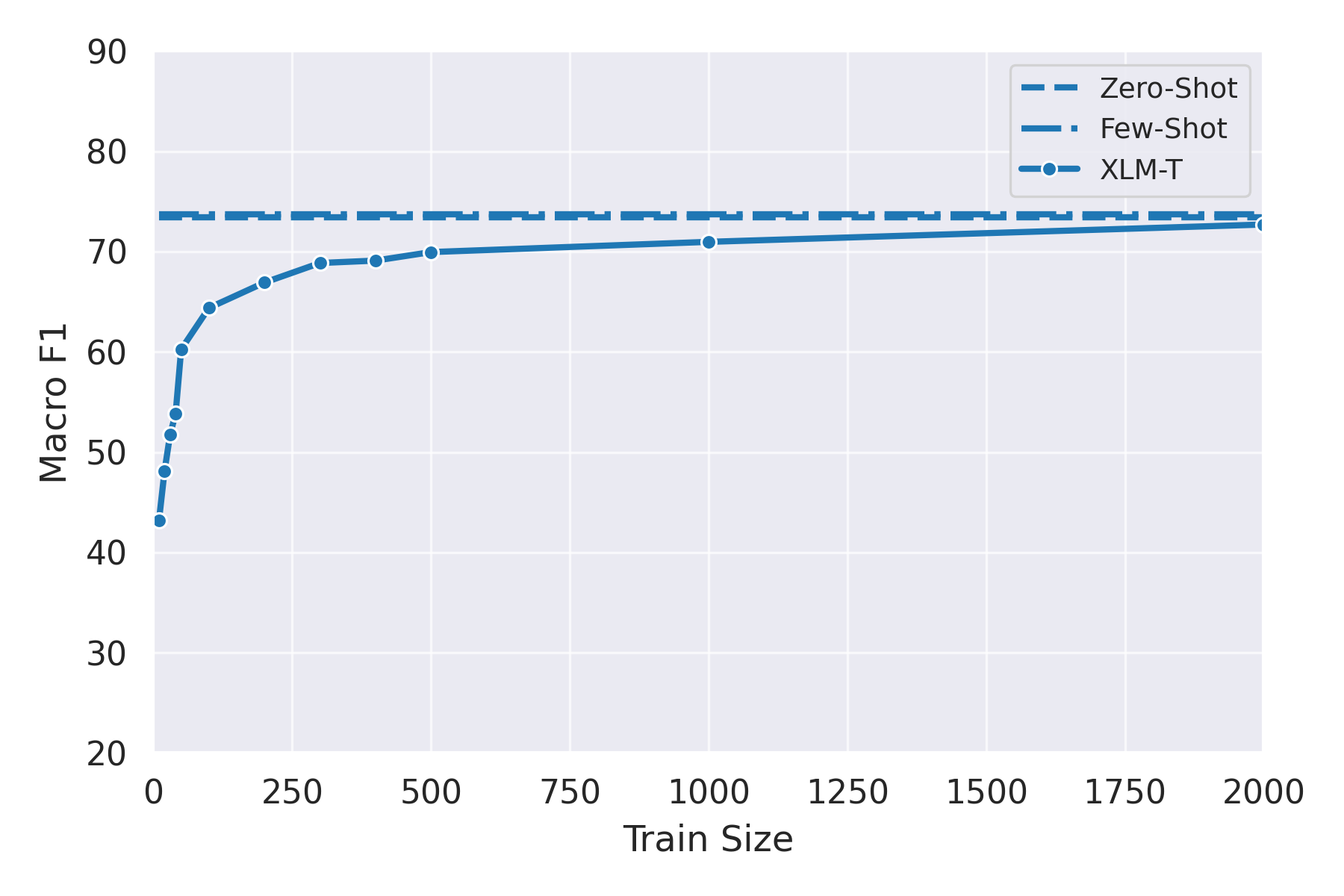}}
    \subfigure[Hindi]{\includegraphics[width=0.23\textwidth]{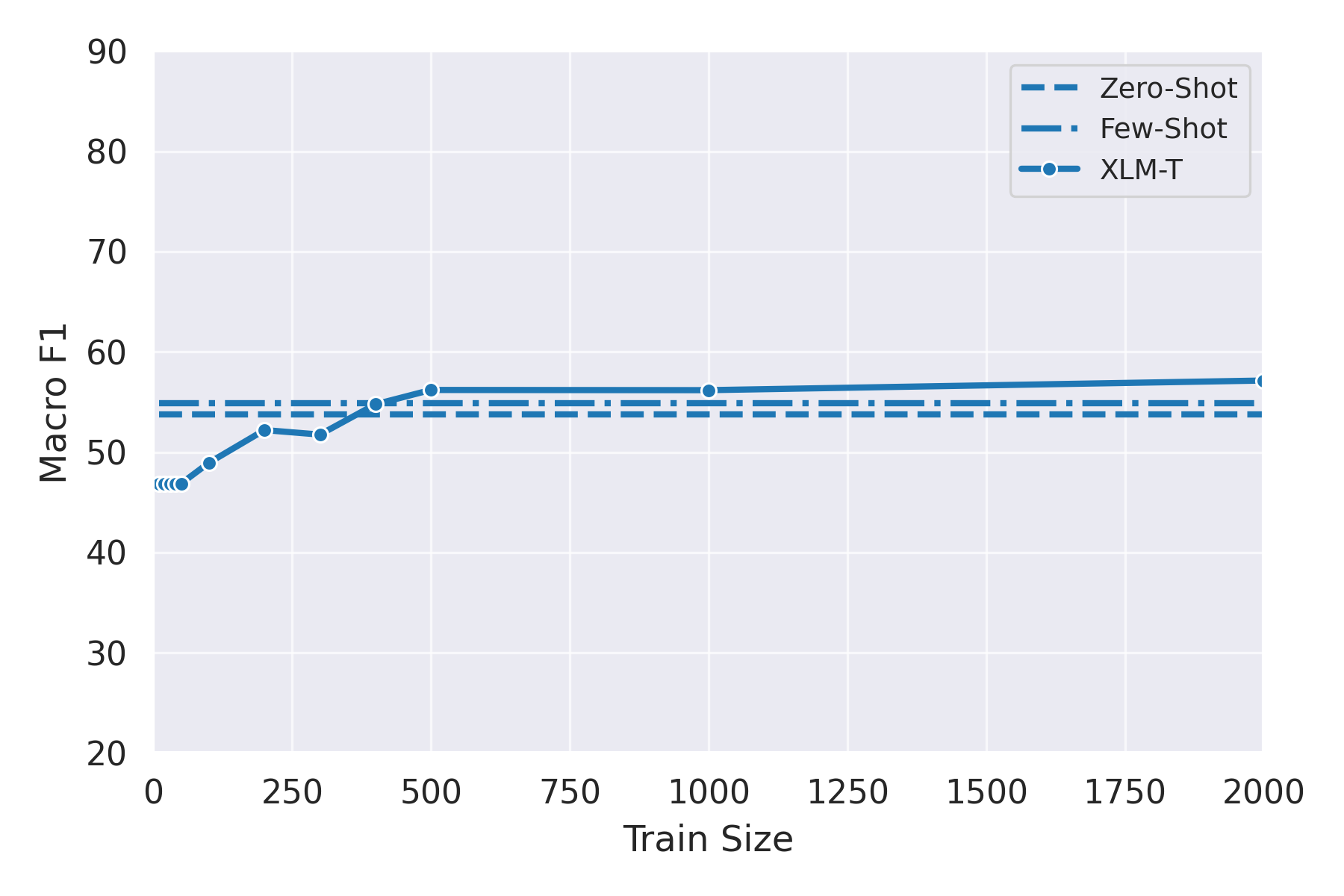}}
    \subfigure[Arabic]{\includegraphics[width=0.23\textwidth]{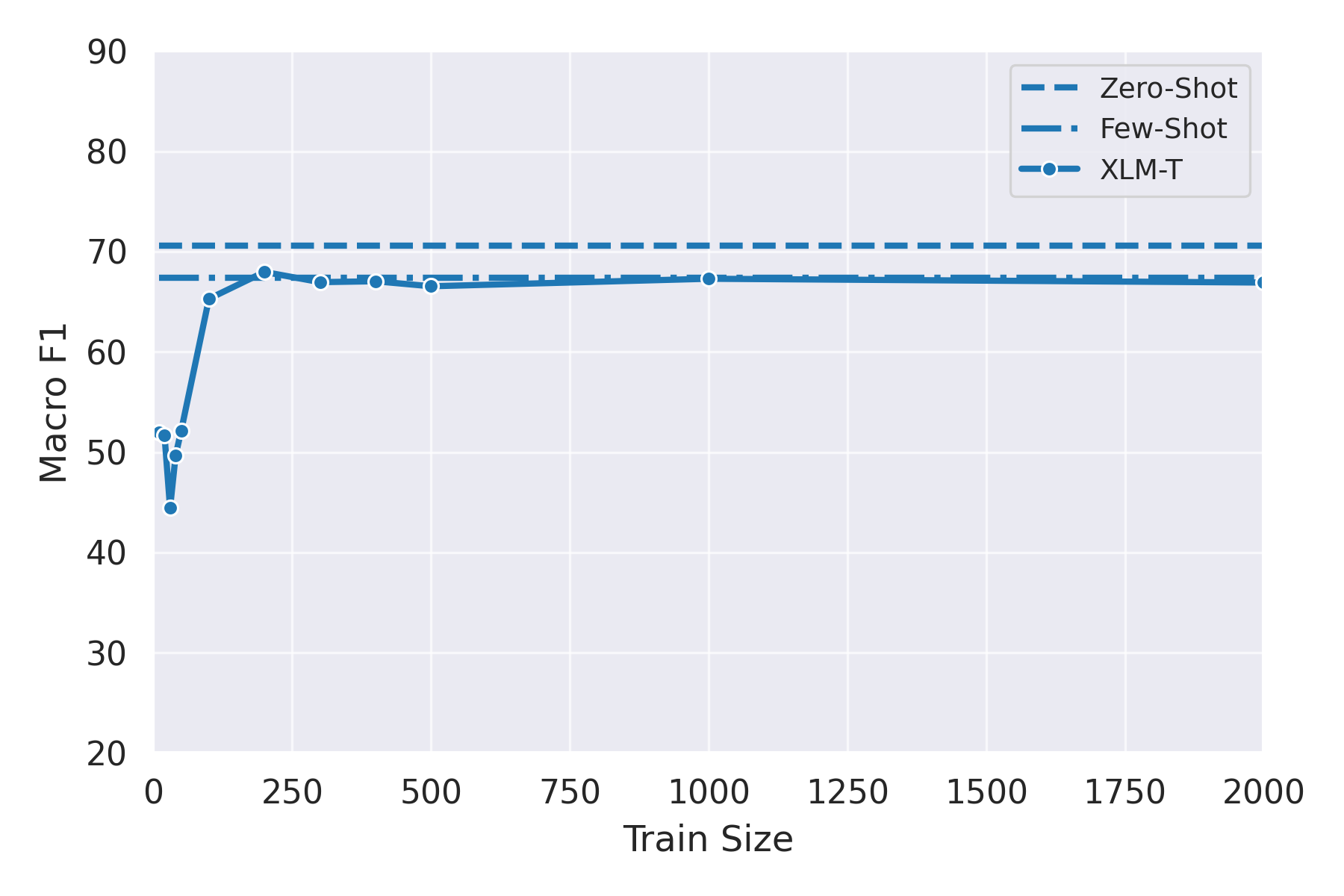}}
    \subfigure[French]{\includegraphics[width=0.23\textwidth]{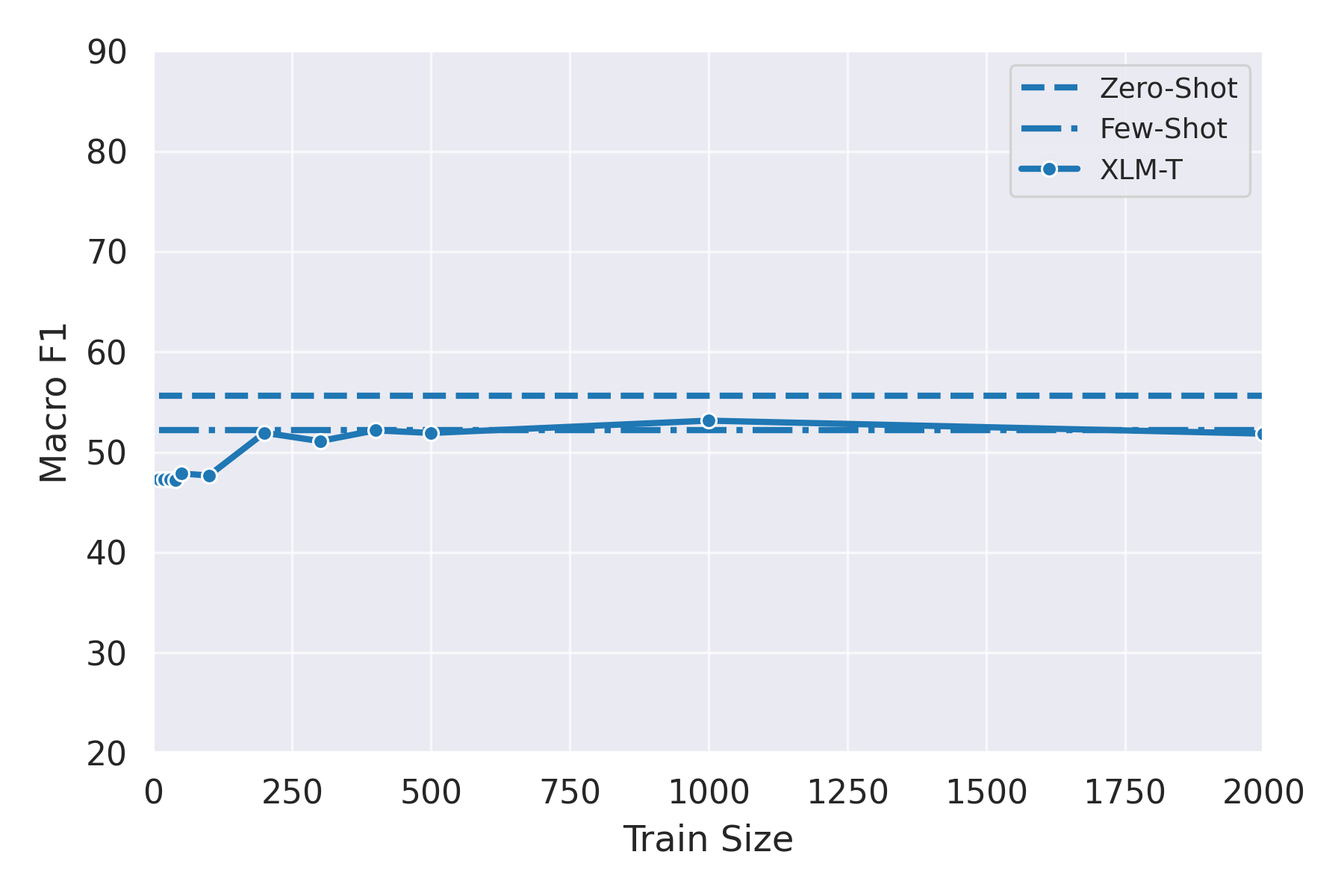}}
    \subfigure[Italian]{\includegraphics[width=0.23\textwidth]{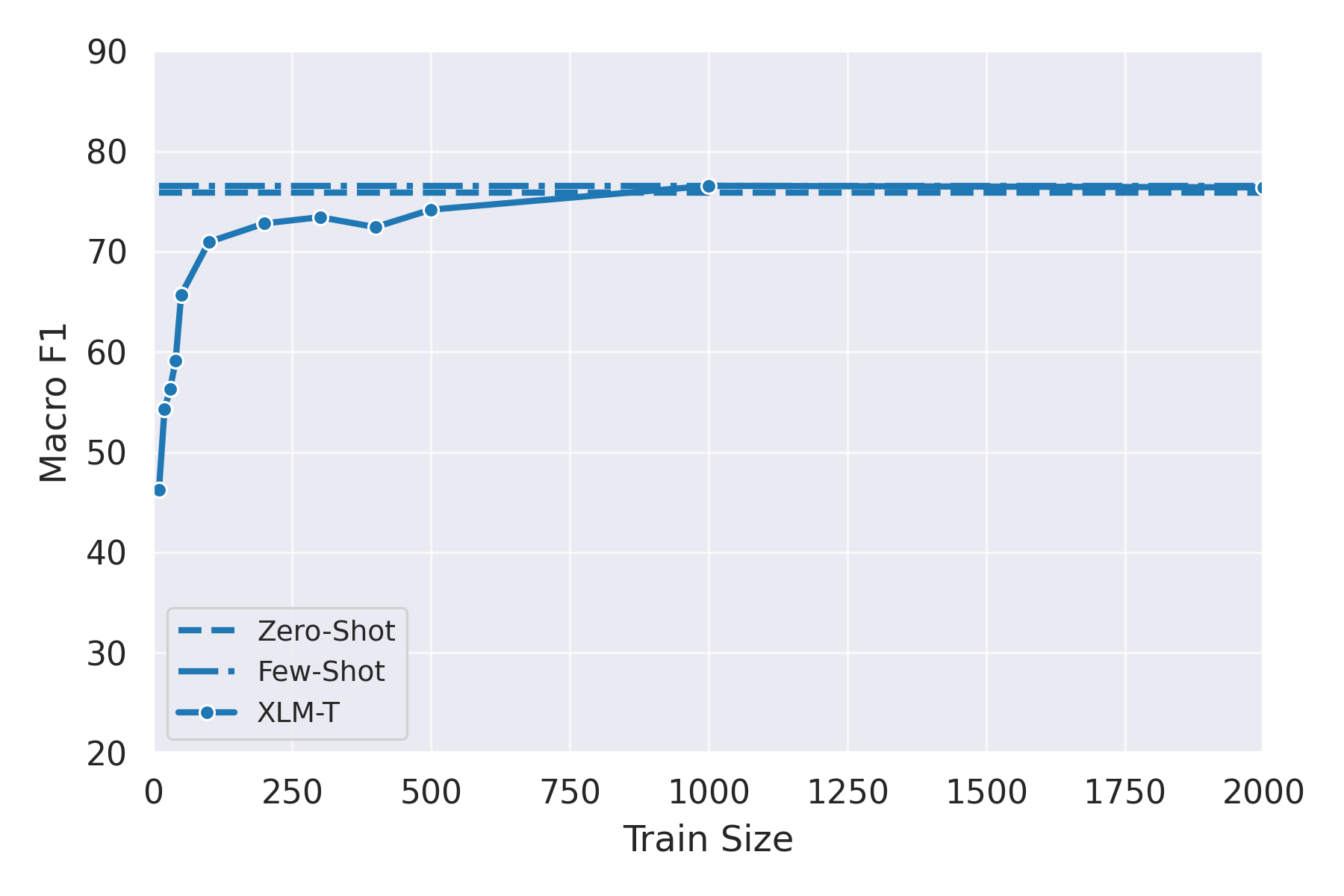}}
    \subfigure[German]{\includegraphics[width=0.23\textwidth]{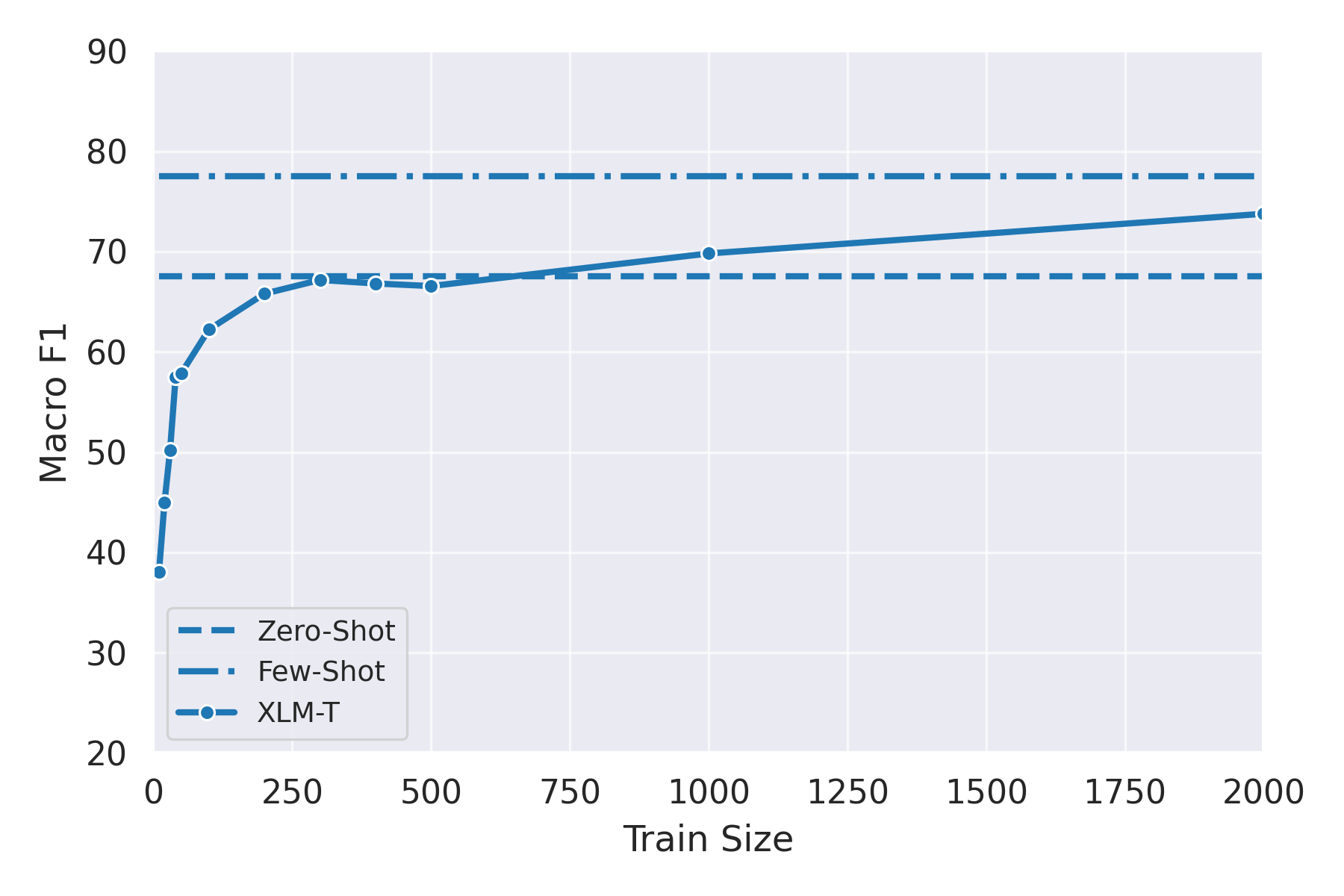}}
    \subfigure[Turkish]{\includegraphics[width=0.23\textwidth]{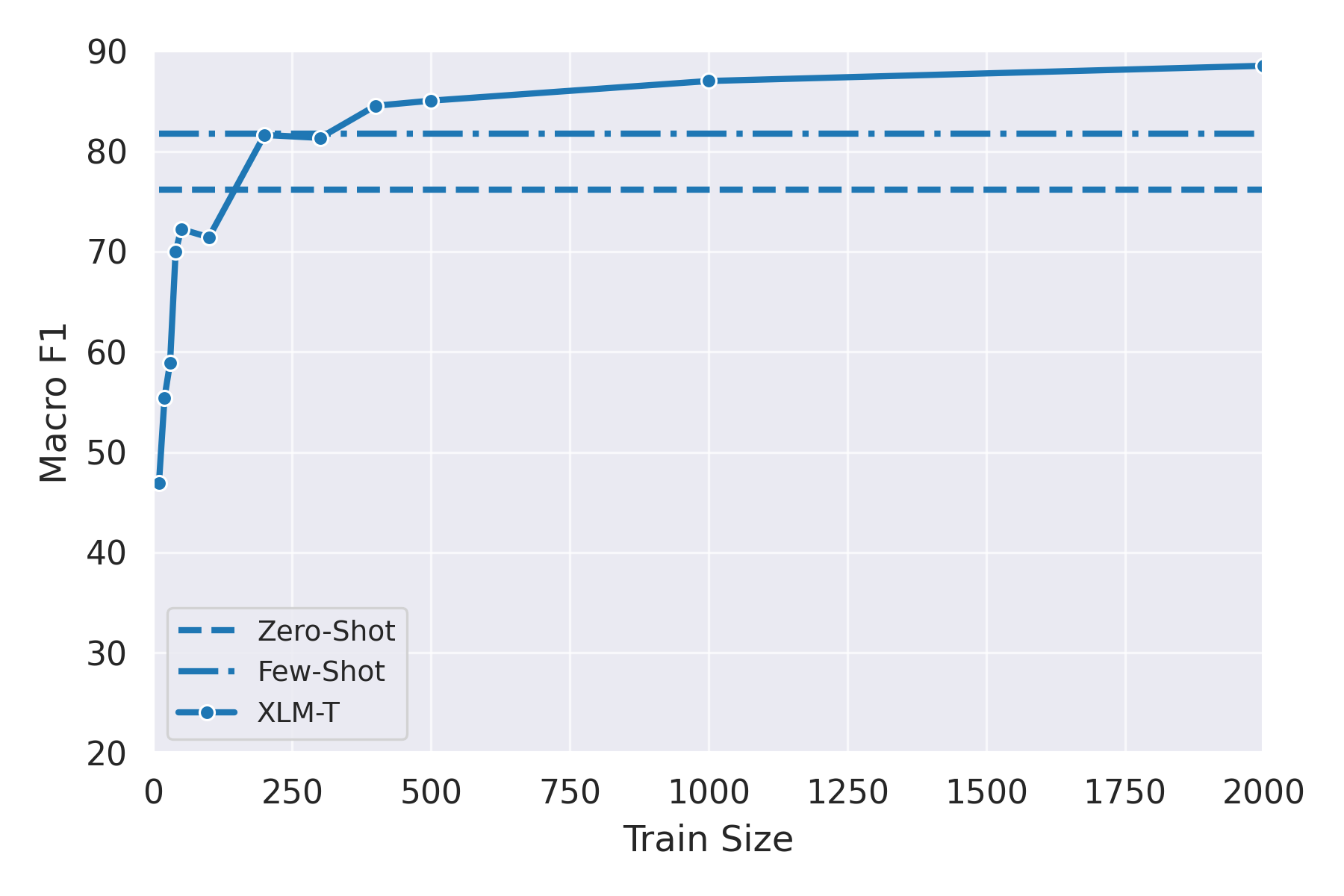}}
    \captionsetup{skip=3pt}  
    \caption{Performance of zero-/few-shot prompted LLMs vs. fine-tuned XLM-T across varying training sizes.}
    \label{fig:all_plots}
\end{figure*}

Figure~\ref{fig:all_plots} illustrates the performance of the XLM-T model fine-tuned on training sets ranging from 10 to 2,000 instances across various languages, alongside the best zero-/few-shot results from instruction-tuned LLMs. Notably, in Portuguese, Arabic, French, and Italian, zero- or few-shot prompting matches or exceeds the performance of XLM-T even when trained on 2,000 labeled examples. In other languages, prompting performs competitively when training data is limited, offering a strong alternative in low-resource settings. As expected, fine-tuning generally surpasses prompting when sufficient labeled data is available, highlighting a practical trade-off between data availability and model adaptation strategy.

\section{Full Prompting Results}
\label{sec:full_prompting_results}
The complete prompting results of the four instruction-tuned LLMs for Spanish and Portuguese are shown in Tables~\ref{tab:spanish} and~\ref{tab:portuguese}.

\begin{table*}[!htp]\centering
\scriptsize
\begin{tabular}{lrrrrrrrrrr}\toprule
& &\multicolumn{4}{c}{Real-world test sets} &\multicolumn{4}{c}{Functional test sets} \\\cmidrule{3-10}
& &llama3 &aya101 &bloomz &qwan &llama3 &aya101 &bloomz &qwan \\\midrule
general & &40.14 &62.56 &44.47 &62.26 &\textbf{86.37} &67.33 &36.30 &\textbf{84.39} \\
classification & &\textbf{63.13} &60.16 &\textbf{54.50} &62.64 &83.92 &39.06 &38.53 &81.16 \\
definition & &62.21 &\textbf{63.68} &55.98 &42.35 &82.79 &58.71 &\textbf{64.88} &80.46 \\
CoT & &50.65 &37.19 &50.26 &42.58 &33.59 &28.08 &60.44 &55.58 \\
multilingual & &57.54 &60.62 &48.66 &63.08 &69.72 &59.04 &29.33 &48.79 \\
NLI & &47.68 &35.74 &58.66 &18.50 &25.18 &37.76 &57.90 &28.44 \\
role-play & &58.99 &60.68 &55.90 &43.79 &78.03 &44.25 &56.81 &55.39 \\
multilingual + definition & &57.54 &60.65 &54.15 &42.43 &76.73 &51.08 &53.11 &47.35 \\
role-play + CoT & &55.42 &26.90 &29.58 &59.92 &74.25 &41.94 &41.27 &73.87 \\
multilingual + CoT & &59.06 &47.15 &30.61 &60.57 &61.37 &42.20 &41.27 &72.68 \\
target & &41.03 &37.41 &36.82 &\textbf{64.77} &47.91 &30.61 &23.38 &81.28 \\
contextual & &60.83 &40.49 &46.25 &61.74 &80.34 &42.69 &44.90 &82.52 \\
translate & &55.91 &39.05 &35.89 &\textbf{64.79} &71.84 &44.42 &40.50 &76.16 \\
distinction & &62.80 &62.94 &36.80 &63.42 &78.06 &\textbf{73.19} &23.53 &78.38 \\\midrule
\multirow{2}{*}{few-shot} &1 &44.76 &23.42 &44.76 &42.11 &86.44 &28.46 &34.41 &53.74 \\
&5 &47.85 &None &50.78 &45.33 &58.63 &None &58.81 &56.90 \\\midrule
\multirow{2}{*}{few shot + CoT} &1 &64.44 &28.09 &54.27 &64.34 &82.93 &32.59 &35.54 &80.68 \\
&5 &\textbf{68.89} &None &55.15 &\textbf{68.90} &\textbf{86.45} &None &38.36 &84.03 \\\midrule
\multirow{2}{*}{few shot + role-play} &1 &43.61 &37.55 &57.06 &66.44 &55.86 &37.88 &36.83 &81.60 \\
&5 &46.01 &None &53.25 &45.88 &56.29 &None &29.77 &83.45 \\\midrule
\multirow{2}{*}{few shot + multilingual} &1 &65.25 &42.51 &\textbf{58.71} &64.41 &84.89 &42.88 &43.10 &81.03 \\
&5 &\textbf{68.35} &None &57.82 &45.28 &84.94 &None &45.03 &82.99 \\\midrule
\multirow{2}{*}{few shot + definition} &1 &64.25 &41.09 &52.40 &64.83 &82.53 &41.99 &29.18 &80.06 \\
&5 &66.85 &None &48.85 &66.94 &83.59 &None &25.00 &\textbf{86.43} \\
\bottomrule
\end{tabular}
\caption{Complete Zero- and Few-shot Prompting Results for \underline{Spanish}.}\label{tab:spanish}
\end{table*}

\begin{table*}[!htp]\centering
\scriptsize
\begin{tabular}{lrrrrrrrrrr}\toprule
& &\multicolumn{4}{c}{Real-world test sets} &\multicolumn{4}{c}{Functional test sets} \\\cmidrule{3-10}
& &llama3 &aya101 &bloomz &qwan &llama3 &aya101 &bloomz &qwan \\\midrule
general & &44.98 &71.08 &45.00 &42.72 &82.71 &64.93 &42.98 &56.25 \\
classification & &67.05 &45.70 &39.12 &70.63 &\textbf{83.37} &34.35 &28.83 &79.50 \\
definition & &45.83 &\textbf{71.51} &\textbf{63.92} &66.00 &79.25 &55.36 &\textbf{66.04} &79.50 \\
CoT & &50.22 &41.54 &56.89 &36.01 &48.04 &28.44 &61.43 &\textbf{82.15} \\
multilingual & &67.64 &67.68 &49.39 &66.97 &76.88 &56.38 &37.26 &75.78 \\
NLI & &49.54 &41.36 &53.68 &8.51 &35.47 &40.45 &58.81 &37.60 \\
role-play & &\textbf{70.79} &65.33 &58.76 &41.59 &\textbf{82.22} &43.78 &56.03 &55.16 \\
multilingual\_definition & &63.29 &60.29 &48.28 &69.49 &69.82 &48.01 &50.48 &78.71 \\
role-play + CoT & &67.14 &40.57 &24.01 &\textbf{73.44} &72.82 &42.30 &41.15 &73.05 \\
multilingual + CoT & &45.91 &55.23 &35.90 &72.06 &78.14 &41.91 &41.25 &73.22 \\
target & &44.27 &38.33 &40.62 &67.79 &46.38 &27.97 &24.57 &80.58 \\
contextual & &66.05 &46.03 &30.90 &57.51 &81.22 &40.76 &52.24 &79.48 \\
translate & &69.40 &63.96 &43.74 &60.41 &76.44 &45.83 &41.49 &75.05 \\
distinction & &69.82 &67.11 &40.62 &59.75 &78.30 &\textbf{72.39} &25.36 &77.36 \\\midrule
\multirow{3}{*}{few shot} &1 &46.16 &21.18 &49.97 &70.60 &85.80 &28.94 &45.59 &51.28 \\
&3 &46.13 &21.07 &55.92 &70.19 &\textbf{86.59} &29.30 &65.08 &53.12 \\
&5 &47.00 &21.19 &59.69 &70.39 &57.48 &29.44 &\textbf{68.70} &53.59 \\\midrule
\multirow{3}{*}{few shot + CoT} &1 &70.63 &26.28 &53.76 &72.55 &83.78 &27.66 &33.75 &78.71 \\
&3 &72.35 &28.78 &55.27 &72.34 &84.25 &26.09 &42.63 &79.99 \\
&5 &\textbf{72.98} &25.80 &55.23 &72.53 &84.22 &22.55 &44.59 &80.57 \\\midrule
\multirow{3}{*}{few shot + role-play} &1 &46.35 &32.91 &55.36 &70.74 &56.07 &30.08 &40.13 &53.55 \\
&3 &69.63 &33.24 &54.29 &72.36 &83.32 &31.57 &38.86 &82.38 \\
&5 &69.78 &31.40 &52.23 &\textbf{72.56} &83.82 &31.31 &35.93 &83.29 \\\midrule
\multirow{3}{*}{few shot + multilingual} &1 &71.81 &44.30 &60.42 &70.64 &85.14 &42.58 &49.75 &80.05 \\
&3 &72.24 &\textbf{44.60} &\textbf{60.78} &70.47 &84.72 &42.70 &55.90 &80.84 \\
&5 &\textbf{73.70} &43.88 &60.54 &71.13 &84.78 &42.30 &55.55 &82.16 \\\midrule
\multirow{3}{*}{few shot + definition} &1 &71.06 &33.39 &50.79 &71.74 &83.19 &45.67 &37.20 &80.40 \\
&3 &70.55 &32.86 &52.60 &71.42 &84.42 &33.07 &37.52 &82.82 \\
&5 &71.42 &32.41 &51.50 &47.59 &84.72 &32.54 &36.51 &\textbf{84.08} \\
\bottomrule
\end{tabular}
\caption{Complete Zero- and Few-shot Prompting Results for \underline{Portuguese}.}\label{tab:portuguese}
\end{table*}
\end{document}